\documentclass{article} %
\pdfoutput=1
\usepackage{iclr2017_conference,times}
\usepackage{url}

\usepackage[utf8]{inputenc} %
\usepackage[T1]{fontenc}    %
\usepackage{url}            %
\usepackage{booktabs}       %
\usepackage{amsfonts}       %
\usepackage{nicefrac}       %
\usepackage{microtype}      %
\usepackage{graphicx}
\usepackage{amsmath}
\usepackage{amssymb}
\usepackage{color}
\usepackage{caption}
\usepackage{subcaption}

\usepackage[colorlinks,linkcolor={red!80!black},citecolor={blue!70!black},urlcolor={blue!80!black}]{hyperref} %

\usepackage{todonotes}

\usepackage[capitalize,noabbrev]{cleveref}

\DeclareMathOperator{\E}{\mathbb{E}}

\newcommand{\distconv}{\stackrel{D}{\to}}
\newcommand{\simiid}{\stackrel{\mathit{iid}}{\sim}}
\newcommand{\acro}[1]{\textsc{\MakeLowercase{#1}}}
\newcommand{\tp}{^\mathsf{T}}

\DeclareMathOperator{\mmd}{\acro{mmd}}
\DeclareMathOperator{\mmdsq}{\acro{mmd}^2}
\DeclareMathOperator{\mmdsqu}{\widehat{\acro{mmd}}_U^2}

\newcommand{\httpurl}[1]{\href{http://#1}{\nolinkurl{#1}}}
\newcommand{\httpsurl}[1]{\href{https://#1}{\nolinkurl{#1}}}

\title{Generative Models and Model Criticism via Optimized Maximum Mean Discrepancy}

\author{%
    Danica J. Sutherland\footnotemark[1]\,\,\,\footnotemark[2]\hfill\qquad\hfill
    Hsiao-Yu Tung\footnotemark[2]\hfill\qquad\hfill
    Heiko Strathmann\footnotemark[1]\hfill\qquad\hfill
    Soumyajit De\footnotemark[1]
    \hfill\null \\
    \textbf{Aaditya Ramdas\footnotemark[3]}\hfill
    \textbf{Alex Smola\footnotemark[2]}\hfill
    \textbf{Arthur Gretton\footnotemark[1]}
    \hspace*{\fill}
    \\
    \footnotemark[1]\; Gatsby Computational Neuroscience Unit, University College London \\
    \footnotemark[2]\; School of Computer Science, Carnegie Mellon University \\
    \footnotemark[3]\; Departments of EECS and Statistics, University of California at Berkeley \\
    \texttt{djs@djsutherland.ml} \; \texttt{htung@cs.cmu.edu}
}

\iclrfinalcopy %

\begin{document}

\maketitle

\begin{abstract}
  We propose a method to optimize the representation and distinguishability of samples from two probability distributions, by maximizing the estimated power of a statistical test based on the maximum mean discrepancy (\acro{mmd}).
  This optimized \acro{mmd} is applied to the setting of unsupervised learning by generative adversarial networks (\acro{GAN}), in which a model attempts to generate realistic samples, and a discriminator attempts to tell these apart from data samples.  In this context, the \acro{MMD} may be used in two roles: first, as a discriminator, either directly on the samples, or on features of the samples. Second, the \acro{mmd} can be used to evaluate the performance of a generative model, by testing the model's samples against a reference data set. In the latter role, the optimized \acro{mmd} is particularly helpful, as it gives an interpretable indication of how the model and data distributions differ, even in cases where individual model  samples are not easily distinguished either by eye or by classifier.

  \emph{This post-publication revision corrects some errors in constants of the estimator \eqref{eq:vhat-m}. The appendix deriving the estimator has been replaced by \citet{unbiased-var-ests}.}
\end{abstract}

\section{Introduction}

Many problems in testing and learning require evaluating distribution
similarity in high dimensions, and on structured data such as images
or audio. When a complex generative model is learned, it is necessary
to provide feedback on the quality of the samples produced. The generative
adversarial network \citep{gan,Gutmann2014} is a popular method for training generative
models, where a rival discriminator attempts to distinguish model
samples from reference data. Training of the generator and discriminator
is interleaved, such that a saddle point is eventually reached in
the joint loss.

A useful insight into the behavior of \acro{GAN}s is to note that when
the discriminator is properly trained, the generator is tasked with minimizing
the Jensen-Shannon divergence measure between the model and data distributions.
When the model is insufficiently powerful to perfectly simulate the test data,
as in most nontrivial settings,
the choice of divergence measure is especially crucial:
it determines which compromises will be made.
A range of adversarial divergences were proposed by \cite{huszar16}, using
a weight to interpolate between \acro{KL}, inverse \acro{KL}, and Jensen-Shannon.
This weight may be interpreted as a prior probability of observing
samples from the model or the real world: when there is a greater
probability of model samples, we approach reverse \acro{KL} and the model
seeks out modes of the data distribution. When there is a greater
probability of drawing from the data distribution, the model approaches
the \acro{KL} divergence, and tries to cover the full support of the data,
at the expense of producing some samples in low probability regions.

This insight was further developed by \cite{NowCseTom16}, who showed
that a much broader range of $f$-divergences can be learned for the
discriminator in adversarial models, based on the variational formulation
of $f$-divergences of \cite{NguWaiJor08}. For a given $f$-divergence,
the model learns the composition of the density ratio (of data to
model density) with the derivative of $f$, by comparing generator
and data samples. This provides a lower bound on the ``true'' divergence
that would be obtained if the density ratio were perfectly known.
In the event that the model is in a smaller class than the true data
distribution, this broader family of divergences implements a variety
of different approximations: some focus on individual modes of the
true sample density, others try to cover the support. It is straightforward
to visualize these properties in one or two dimensions \citep[Figure 5]{NowCseTom16},
but in higher dimensions it becomes difficult to anticipate or
visualize the behavior of these various divergences.

An alternative family of divergences
are the integral probability metrics \citep{Mueller97}, which find a witness function to distinguish samples from $P$ and $Q$.\footnote{Only the total variation distance is both an $f$-divergence and an \acro{ipm} \citep{SriFukGreSchetal12}.}
A popular such class of witness functions in \acro{GAN}s is the maximum mean
discrepancy \citep{mmd-jmlr}, simultaneously proposed by \citet{dziugaite:mmd-gan} and \citet{li:gmmn}.
The architecture used in these two approaches is actually quite different:
Dziugaite et al. use the \acro{MMD} as a discriminator directly at the level
of the generated and test images,
whereas Li et al. apply the \acro{MMD} on input features
learned from an autoencoder, and share the decoding layers of the
autoencoder with the generator network (see their Figure 1(b)). The
generated samples have better visual quality in the latter method, but it
becomes difficult to analyze and interpret the algorithm given the
interplay between the generator and discriminator networks.
In a related approach,  \cite{SalGooZarCheetal16} propose to use
feature matching,
where the generator is tasked with minimizing the squared distance between expected
discriminator features under the model and data distributions, thus retaining
the adversarial setting.

In light of these varied approaches to discriminator training, it is important
to be able to evaluate quality of samples
from a generator against reference data.
An approach used in several studies
is to obtain a Parzen window estimate of the density and compute the log-likelhiood  \citep{gan,NowCseTom16,BreBenVin11}.
Unfortunately, density estimates in such high dimensions
are known to be very unreliable both in theory \citep[Ch. 6]{Wasserman06}
and in practice \citep{note-on-generative-models}.
We can instead ask humans to evaluate
the generated images \citep{NIPS2015_5773,SalGooZarCheetal16}, but while evaluators
should be able to distinguish cases where the samples are over-dispersed
(support of the model is too large), it may be more difficult to find
under-dispersed samples (too concentrated at the modes), or imbalances in the proportions of different shapes,
since the samples themselves
will be plausible images. Recall that different divergence measures
result in different degrees of mode-seeking: if we rely
on human evaluation, we may tend towards always using divergences
with under-dispersed samples.

 We propose to use the \acro{MMD}  to distinguish generator and reference data, with features and kernels
 chosen to
 maximize the
test power of the quadratic-time \acro{MMD} of \cite{mmd-jmlr}.
Optimizing \acro{MMD} test power requires a  sophisticated treatment due to the different form
of the null and alternative distributions (\cref{sec:mmd}). We
also develop an efficient approach to obtaining quantiles of the \acro{MMD} distribution under the null
(\cref{sec:permutation}). We demonstrate on simple artificial data that simply maximizing
the \acro{MMD}  \citep[as in][]{max-mmd}   provides a less powerful test than our approach of explicitly
maximizing test power. Our procedure applies even when our definition
of the \acro{MMD} is computed on features of the inputs,
since these can also be trained by power maximization.

When designing an optimized \acro{MMD} test,
we should choose a kernel family that allows us to visualize
where the probability mass of the two samples differs most.
In our experiments on \acro{gan} performance evaluation, we use an automatic relevance determination (\acro{ARD}) kernel over the output
dimensions, and learn which coordinates differ meaningfully by finding
which kernels retain significant bandwidth when the test power is
optimized.
We may further apply the method of \citet[][Section 5]{LloGha15} to visualize the witness function
associated with this \acro{MMD},
 by finding those model
and data samples occurring at the maxima and minima of the witness
function (i.e., the samples from one distribution least likely to
be in high probability regions of the other).
The optimized witness function gives a test with greater power than
a standard \acro{rbf} kernel, suggesting that the associated witness function peaks are an
improved representation of where the distributions differ.
We also propose a novel generative model based on the feature matching idea of \cite{SalGooZarCheetal16},
using \acro{MMD} rather than their ``minibatch discrimination'' heuristic, for a more principled and more stable
enforcement of sample diversity, without requiring labeled data.

\section{Maximizing Test Power of a Quadratic MMD Test} \label{sec:mmd}

Our methods rely on optimizing the power of a two-sample test over the choice of kernel.
We first describe how to do this,
then review alternative kernel selection approaches.

\subsection{MMD and test power}\label{sec:MMDandTestPower}

We will begin by reviewing the maximum mean discrepancy and its use in two-sample tests.
Let $k$ be the kernel of a reproducing kernel Hilbert space (\acro{rkhs}) $\mathcal{H}_k$ of functions on a set $\mathcal X$.
We assume that $k$ is measurable and bounded,
$\sup_{x \in \mathcal X} k(x, x) < \infty$.
Then the \acro{mmd} in $\mathcal{H}_k$ between two distributions $P$ and $Q$ over $\mathcal X$ is \citep{mmd-jmlr}:
\begin{equation}
  \acro{mmd}_k^2(P, Q)
  := \E_{x, x'}\left[ k(x, x') \right]
  + \E_{y, y'}\left[ k(y, y') \right]
  - 2 \E_{x, y}\left[ k(x, y) \right]
\label{eq:pop-mmd}
\end{equation}
where $x, x' \simiid P$ and $y, y' \simiid Q$.
Many kernels, including the popular Gaussian \acro{rbf}, are \emph{characteristic} \citep{kernel-conditional-dep,SriGreFukLanetal10},
which implies that the \acro{MMD} is a metric,
and in particular that $\mmd_k(P, Q) = 0$ if and only if $P = Q$,
so that tests with any characteristic kernel are consistent.
That said, different characteristic kernels will yield different test powers for finite sample sizes,
and so we wish to choose a kernel $k$ to maximize the test power.
Below, we will usually suppress explicit dependence on $k$.

Given $X = \{X_1, \dots, X_m \} \simiid P$ and $Y = \{ Y_1, \dots, Y_m \} \simiid Q$,\footnote{We assume for simplicity that the number of samples from the two distributions is equal.}
one estimator of $\mmd(P, Q)$ is
\begin{equation}
  \mmdsqu(X, Y)
  := \frac{1}{\binom{m}{2}} \sum_{i \ne i'} k(X_i, X_{i'})
  + \frac{1}{\binom{m}{2}} \sum_{j \ne j'} k(Y_j, Y_{j'})
  - \frac{2}{\binom{m}{2}} \sum_{i \ne j} k(X_i, Y_j)
\label{eq:mmd-u}
.\end{equation}
This estimator is unbiased, and has nearly minimal variance among unbiased estimators \citep[Lemma~6]{mmd-jmlr}.

Following \citet{mmd-jmlr},
we will conduct a hypothesis test with null hypothesis $H_0 : P = Q$
and alternative $H_1 : P \ne Q$,
using test statistic $m \mmdsqu(X, Y)$.
For a given allowable probability of false rejection $\alpha$,
we choose a test threshold $c_\alpha$ and reject $H_0$ if $m \mmdsqu(X, Y) > c_\alpha$.

Under $H_0 : P = Q$,
 $m \mmdsqu(X, Y)$  converges asymptotically to a distribution that depends on the unknown distribution $P$ \citep[Theorem~12]{mmd-jmlr};
we thus cannot evaluate the test threshold $c_\alpha$ in closed form.
We instead estimate a data-dependent threshold $\hat c_\alpha$ via permutation:
randomly partition the data $X \cup Y$ into $X'$ and $Y'$ many times,
evaluate $m \mmdsqu(X', Y')$ on each split,
and estimate the $(1-\alpha)$th quantile $\hat c_\alpha$ from these samples.
\Cref{sec:permutation} discusses efficient computation of this process.

We now describe a mechanism to choose the kernel $k$ so as to maximize the power of its associated test.
First, note that under the alternative $H_1 : P \ne Q$,
$\mmdsqu$ is asymptotically normal,
\begin{equation}
  \frac{\mmdsqu(X, Y) - \mmdsq(P, Q)}{\sqrt{V_m(P, Q)}}
  \distconv
  \mathcal{N}(0, 1)
\label{eq:mmdu-alt}
,\end{equation}
where
$V_m(P, Q)$ denotes the asymptotic variance of the $\mmdsqu$ estimator for samples of size $m$ from $P$ and $Q$ \citep{serfling:approx-thms}.
The power of our test is thus,
using ${\Pr}_1$ to denote probability under $H_1$,
\begin{align}
  {\Pr}_1\left( m \mmdsqu(X, Y) > \hat c_\alpha \right)
  &= {\Pr}_1\left(
    \frac{\mmdsqu(X, Y) - \mmdsq(P, Q)}{\sqrt{V_m(P, Q)}}
    > \frac{\hat c_\alpha / m - \mmdsq(P, Q)}{\sqrt{V_m(P, Q)}}
  \right)
\notag
\\&\to \Phi\left( \frac{\mmdsq(P, Q)}{\sqrt{V_m(P, Q)}} - \frac{c_\alpha}{m \sqrt{V_m(P, Q)}} \right)
\label{eq:asymp-power}
\end{align}
where $\Phi$ is the \acro{cdf} of the standard normal distribution.
The second step follows by \eqref{eq:mmdu-alt} and the convergence of $\hat c_\alpha \to c_\alpha$ \citep{AlbaFernandez2008}.
Test power is therefore maximized by maximizing the argument of $\Phi$: i.e.
increasing the ratio of $\mmdsq(P, Q)$ to $\sqrt{V_m(P, Q)}$,
and reducing the ratio of $c_\alpha$ to $m \sqrt{V_m(P, Q)}$.

For a given kernel $k$,
$V_m$ is $O( m^{-1} )$,
while both $c_\alpha$ and $\mmd^2$ are constants.
Thus the first term is $O\big( \sqrt m \big)$,
and the second is $O\big( 1 / \sqrt{m} \big)$.
Two situations therefore arise: when $m$ is small relative to the difference in $P$ and $Q$ (i.e.,
we are close to the null), both terms need to be taken into acccount to maximize test power.
Here, we propose to maximize \eqref{eq:asymp-power} using the efficient computation of $\hat c_\alpha$ in \cref{sec:permutation}.
As $m$ grows, however, we can asymptotically maximize the power of the test by choosing a kernel $k$ that maximizes the $t$-statistic
$t_k(P, Q) := \mmd_k^2(P, Q) / \sqrt{V_m^{(k)}(P, Q)}$.
In practice, %
we maximize an estimator of $t_k(P,Q)$ given by $\hat t_k(X, Y) := \mmdsqu(X, Y) / \sqrt{\widehat V_m(X, Y)}$,
with $\widehat V_m(X, Y)$ discussed shortly.

To maintain the validity of the hypothesis test,
we will need to divide the observed data $X$ and $Y$ into a ``training sample,''
used to choose the kernel,
and a ``testing sample,'' used to perform the final hypothesis test with the learned kernel.

We next consider families of kernels over which to optimize.
The most common kernels used for \acro{mmd} tests are standard kernels from the  literature,
e.g.\ the Gaussian \acro{rbf}, Mat\'ern, or Laplacian kernels.
It is the case, however, that for any function $z : \mathcal X_1 \to \mathcal X_2$ and any kernel $\kappa : \mathcal X_2 \times \mathcal X_2 \to \mathbb R$,
the composition $\kappa \circ z$ is also a kernel on $\mathcal X_1$.\footnote{If $z$ is injective and $\kappa$ characteristic, then $\kappa \circ z$ is characteristic. Whether any fixed $\kappa \circ z$ is consistent, however, is less relevant than the power of the $\kappa \circ z$ we choose~---~which is what we maximize.}
We can thus choose a function $z$ to extract meaningful features of the inputs,
and use a standard kernel $\kappa$ to compare those features.
We can select such a function $z$ (as well as $\kappa$)
by performing kernel selection on the family of kernels $\kappa \circ z$.
To do so, we merely need to maximize $\hat t_{\kappa \circ z}(X, Y) $
through standard optimization techniques based on the gradient of $\hat t_{\kappa \circ z}$ with respect to the parameterizations of $z$ and $\kappa$.

We now give an expression for an empirical estimate $\widehat V_m$
of the variance $V_m(P, Q)$ that appears in our test power.
This estimate is similar to that given by \citet[][Appendix A.1]{relative-mmd},
but incorporates second-order terms and corrects some small sources of bias.
Though the expression is somewhat unwieldy,
it is defined by various sums of the kernel matrices
and is differentiable with respect to the kernel $k$.

\begingroup
\newcommand{\Kxy}{K_{XY}}
\newcommand{\Ktxx}{\tilde K_{XX}}
\newcommand{\Ktyy}{\tilde K_{YY}}
\newcommand{\one}{e}
\makeatletter
\DeclareRobustCommand{\norm}{\@ifstar\@@norm\@norm}
\newcommand{\@norm}[1]{\left\lVert #1 \right\rVert}
\newcommand{\@@norm}[1]{\lVert #1 \rVert}
\makeatother

$V_m(P, Q)$ is given in terms of expectations of $k$ under $P$ and $Q$ in \cref{sec:variance}.
We replace these expectations with finite-sample averages,  giving us the required estimator.
Define matrices $\Kxy$, $\Ktxx$, and $\Ktyy$ by
$(\Kxy)_{i,j} = k(X_i, Y_j)$,
$(\Ktxx)_{ii} = 0$, $(\Ktxx)_{ij} = k(X_i, X_j)$ for $i \ne j$,
and $\Ktyy$ similarly to $\Ktxx$.
Let $e$ be an $m$-vector of ones,
and use the falling factorial notation $(m)_r := m (m-1) \cdots (m - r + 1)$.
Then an unbiased estimator for $V_m(P, Q)$ is:
\begin{equation} \label{eq:vhat-m}
\begin{aligned}
\widehat V_m
&:=
    \frac{4}{(m)_4} \left[ \norm{\Ktxx \one}^2 + \norm{\Ktyy \one}^2 \right]
  + \frac{4 (m^2 - m - 1)}{m^3 (m-1)^2} \left[ \norm{\Kxy \one}^2 + \norm*{\Kxy\tp \one}^2 \right]
\\&\qquad
  - \frac{8}{m^2 (m^2 - 3 m + 2)} \left[ \one\tp \Ktxx \Kxy \one + \one\tp \Ktyy \Kxy\tp \one \right]
\\&\qquad
  + \frac{8}{m^2 (m)_3} \left[ \left( \one\tp \Ktxx \one + \one\tp \Ktyy \one \right) \left( \one\tp \Kxy \one \right) \right]
\\&\qquad
  - \frac{2 (2 m - 3)}{(m)_2 (m)_4} \left[
      \left(\one\tp \Ktxx \one \right)^2
    + \left(\one\tp \Ktyy \one \right)^2
  \right]
  - \frac{4 (2 m - 3)}{m^3 (m-1)^3} \left[
      \left(\one\tp \Kxy \one\right)^2
  \right]
\\&\qquad
  - \frac{2}{m (m^3 - 6 m^2 + 11 m - 6)} \left[ \norm{\Ktxx}_F^2 + \norm{\Ktyy}_F^2 \right]
  + \frac{4 (m-2)}{m^2 (m-1)^3} \norm{\Kxy}_F^2
.\end{aligned}
\end{equation}
\endgroup

\subsection{Other Approaches to MMD Kernel Selection } \label{sec:mmd-related}
The most common practice in performing two-sample tests with \acro{mmd} is to use a Gaussian \acro{rbf} kernel,
with bandwidth set to the median pairwise distance among the joint data.
This heuristic often works well,
but fails when the scale on which $P$ and $Q$ vary differs from the scale of their overall variation (as in the synthetic experiment of \cref{sec:experiments}).
\citet{RamRedPocSinWas15,reddi:high-dim-mmd} study the power of the median heuristic in high-dimensional problems,
and justify its use for the case where the means of $P$ and $Q$ differ.

An early heuristic for improving test power was
 to simply maximize $\mmdsqu$.
\citet{max-mmd} proved that, for certain classes of kernels, this yields a consistent test.
As further shown by Sriperumbudur et al., however, maximizing \acro{mmd}  amounts to minimizing training {\em classification
  error} under linear loss.  Comparing with \eqref{eq:asymp-power}, this is plainly not
an optimal approach for maximizing {\em test} power, since variance is ignored.\footnote{With regards to classification vs testing: there has been initial work by \citet{classificationTesting}, who study the simplest setting of the two multivariate Gaussians with known covariance matrices. Here, one can use linear classifiers, and the two sample test boils down to testing for differences in means. In this setting, when the classifier is chosen to be Fisher's \acro{LDA}, then using the classifier accuracy on held-out data as a test statistic turns out to be minimax optimal in ``rate'' (dependence on dimensionality and sample size) but not in constants, meaning that there do exist tests which achieve the same power with fewer samples. The result has been proved only for this statistic and setting, however, and generalization to other statistics and settings is an open question.}
One can also consider maximizing criteria based on cross validation \citep{sugiyama:lst-sq-2samp,mmd-ratio-linear,heiko-masters}.
This approach is not differentiable, and thus difficult to maximize among more than a fixed set of candidate kernels.
Moreover, where this cross-validation is used to maximize the \acro{mmd} on a validation set \citep[as in][]{sugiyama:lst-sq-2samp}, it again amounts to maximizing classification performance rather than test performance, and is suboptimal in the latter setting \citep[][Figure 1]{mmd-ratio-linear}.
Finally, \citet{mmd-ratio-linear} previously studied direct optimization of the power of an \acro{mmd} test
for a streaming estimator of the \acro{mmd}, for which optimizing the ratio of the empirical statistic to its variance
also optimizes test power.
This streaming estimator uses data very inefficiently, however,
often requiring $m^2$ samples to achieve power comparable to tests based on $\mmdsqu$ with $m$ samples \citep{RamRedPocSinWas15}.

\section{Efficient Implementation of Permutation Tests for \texorpdfstring{$\mmdsqu$}{the Quadratic MMD}} \label{sec:permutation}
Practical implementations of tests based on $\mmdsqu$ require efficient estimates of the test threshold $\hat c_\alpha$.
There are two known test threshold estimates that lead to a consistent test:
the permutation test mentioned above,
and a more sophisticated null distribution estimate based on approximating the eigenspectrum of the kernel,
previously reported to be faster than the permutation test \citep{mmd-fast-consistent}.
In fact, the relatively slow reported performance of the permutation approach was due to
the naive Matlab implementation of the permutation test in the code accompanying  \citet{mmd-jmlr}, which creates a new copy of the kernel matrix for every permutation.
We show here that, by careful design, permutation thresholds can be computed substantially faster -- even when compared to parallelized state-of-the-art spectral solvers (not used by \citeauthor{mmd-jmlr}).

First, we observe that we can avoid copying the kernel matrix simply by generating permutation
indices for each null sample and accessing the precomputed kernel matrix in permuted order.
In practice, however, this does not give much performance gain due to the random nature of memory-access which
conflicts with how modern \acro{CPU}s implement caching.
Second, if we rather maintain an inverse map of the permutation indices, we can
easily traverse the matrix in a sequential fashion.
This approach exploits the hardware prefetchers and reduces the number of \acro{CPU}
cache misses from almost 100\% to less than 10\%.
Furthermore, the sequential access pattern of the kernel matrix enables us to invoke multiple threads
for computing the null samples, each traversing the matrix sequentially, without compromising the locality of reference in the
\acro{CPU} cache.

We consider an example problem of computing the test using $200$ null distribution samples on $m=2000$ two-dimensional samples,
comparing a Gaussian to a Laplace distribution with matched moments.
We compare our optimized permutation test against a spectral test using the highly-optimized (and proprietary) state-of-the-art spectral solver of Intel's \acro{MKL} library \citep{intel2007intel}.
All results are averaged over 30 runs; the variance across runs was negligible.

\begin{figure}
\centering
\includegraphics[scale=1]{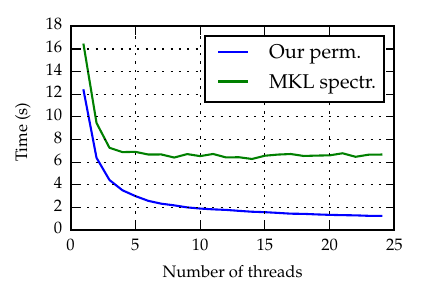} \hspace{-.5cm}
\includegraphics[scale=1]{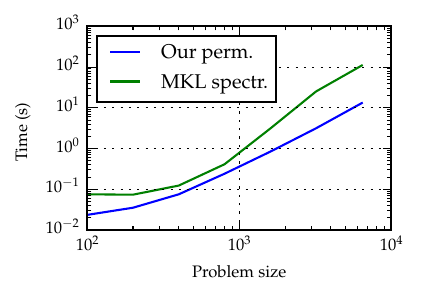}
\caption{Runtime comparison for sampling the null distribution. We compare our optimized permutation approach to the spectral method using Intel's \acro{MKL} spectral solver. Time spent precomputing the kernel matrix is not included. \emph{Left:} Increasing number of threads for fixed problem size $m=2000$. Single-threaded times of other implementations: Matlab reference spectral 381s, Python permutation 182s, Shogun spectral (eigen3) 87s. \emph{Right:} Increasing problem sizes using the maximum number of 24 system threads. }
\label{fig:mmd_perm_performance}
\end{figure}

\cref{fig:mmd_perm_performance} (left) shows the obtained speedups as the number of computing threads grow for $m=2000$.
Our implementation is not only faster on a single thread, but also saturates more slowly as the number of threads increases.
\cref{fig:mmd_perm_performance} (right) shows timings for increasing problem sizes (i.e.\ $m$) when using all available system threads (here 24).
For larger problems, our permutation implementation (scaling as $\mathcal{O}(m^2)$)  is an order of magnitude faster than the spectral test (scaling as $\mathcal{O}(m^3)$).
For smaller problems (for which  \citeauthor{mmd-jmlr} suggested the spectral test), there is still a significant performance increase.

For further reference, we also report timings of available non-parallelized implementations for $m=2000$, compared to our version's {12s} in \cref{fig:mmd_perm_performance} (left): {87s} for an open-sourced spectral test in Shogun using eigen3 \citep{soeren_sonnenburg_2016_51537, eigenweb}, {381s} for the reference Matlab spectral implementation \citep{mmd-jmlr}, and {182s} for a naive Python permutation test that partly avoids copying via masking. (All of these times also exclude kernel computation.)

\section{Experiments} \label{sec:experiments}
Code for these experiments is available at \httpsurl{github.com/djsutherland/opt-mmd}.

\begin{figure}[ht]
  \centering
  \begin{subfigure}[b]{0.21\textwidth}
    \includegraphics[width=\textwidth]{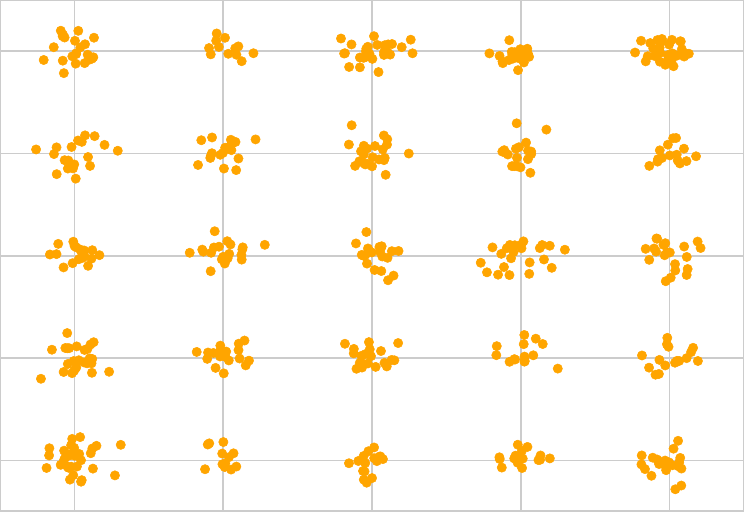}
    \\
    ~\\
    \includegraphics[width=\textwidth]{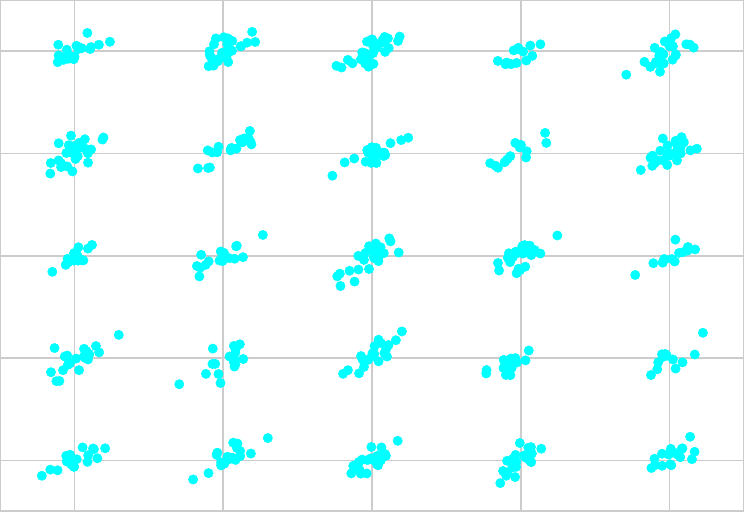}
    \caption{Samples of size $500$ with $\varepsilon = 6$ from $P$ (top) and $Q$ (bottom).}
    \label{fig:blobs:xy}
  \end{subfigure}
  ~
  \begin{subfigure}[b]{0.34\textwidth}
    \includegraphics[width=\textwidth]{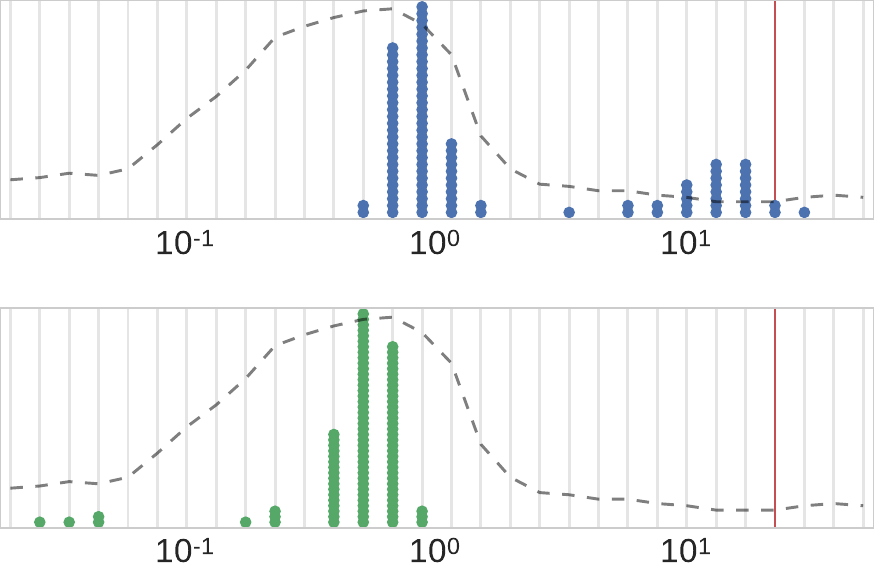}
    \caption{Bandwidths chosen by maximizing $\mmdsqu$ (top, blue) and $\hat t$ (bottom, green), as well as the median heuristic (red), for $\varepsilon = 6$. Gray lines show the power of each bandwidth: $\sigma = 0.67$ had power 96\%, $\sigma = 10$ had 10\%.}
    \label{fig:blobs:choices}
  \end{subfigure}
  ~
  \begin{subfigure}[b]{0.4\textwidth}
    \includegraphics[width=\textwidth]{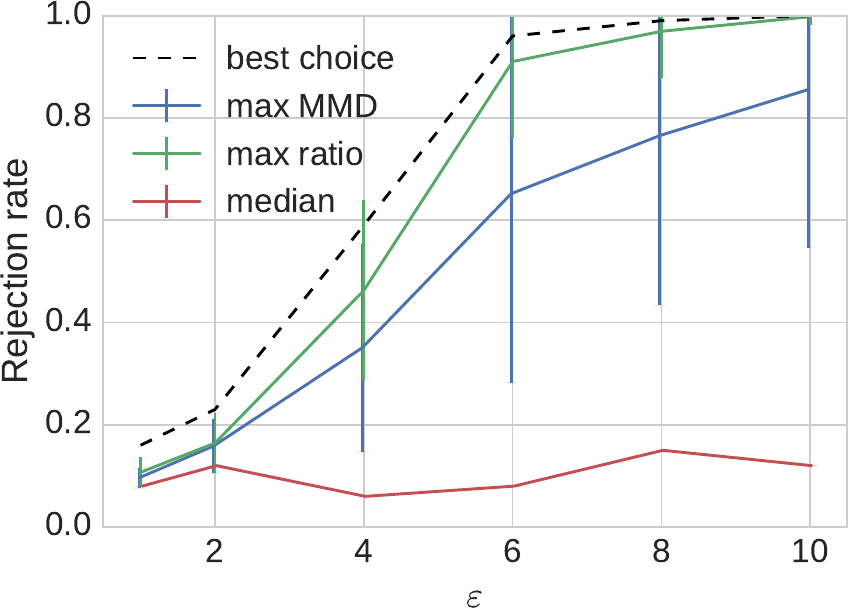}
    \caption{Mean and standard deviation of rejection rate as $\varepsilon$ increases. ``Best choice'' shows the mean power of the bandwidth with highest rejection rate for each problem.}
    \label{fig:blobs:power}
  \end{subfigure}
  \caption{Results for the Blobs problem. Maximizing $\hat t$ performs near-optimally.}
  \label{fig:blobs}
\end{figure}

\paragraph{Synthetic data}
We consider the problem of bandwidth selection for Gaussian \acro{rbf} kernels on the Blobs dataset of \citet{mmd-ratio-linear}.
$P$ here is a $5 \times 5$ grid of two-dimensional standard normals, with spacing 10 between the centers.
$Q$ is laid out identically, but with covariance $\frac{\varepsilon - 1}{\varepsilon + 1}$ between the coordinates (so that the ratio of eigenvalues in the variance is $\varepsilon$.)
\Cref{fig:blobs:xy} shows two samples from $X$ and $Y$ with $\varepsilon = 6$.
Note that when $\varepsilon = 1$, $P = Q$.

For $\varepsilon \in \{1, 2, 4, 6, 8, 10\}$,
we take $m = 500$ samples from each distribution
and compute $\mmdsqu(X, Y)$, $\widehat V_m(X, Y)$, and $\hat c_{0.1}$ using $1\,000$ permutations,
for Gaussian \acro{rbf} kernels with each of 30 bandwidths.
We repeat this process 100 times.
\Cref{fig:blobs:choices} shows that the median heuristic always chooses too large a bandwidth.
When maximizing \acro{mmd} alone, we see a bimodal distribution of bandwidths, with
a significant number of samples falling into the region with low test power.
The variance of $\mmdsqu$ is much higher in this region, however, hence optimizing  the ratio $\hat t$ never returns these bandwidths.
\Cref{fig:blobs:power} shows that maximizing $\hat t$ outperforms maximizing the \acro{mmd} across a variety of problem parameters,
and performs near-optimally.

\begin{figure}
  \centering
  \begin{subfigure}{.2\textwidth}
  \begin{subfigure}[c]{\textwidth}
    \includegraphics[width=\textwidth]{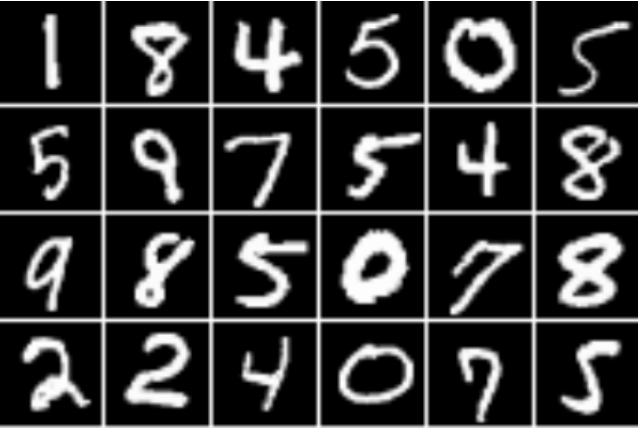}
    \caption{Dataset samples.}
    \label{fig:mnist:true:samp}
  \end{subfigure}
  \\
  \begin{subfigure}[c]{\textwidth}
    \includegraphics[width=\textwidth]{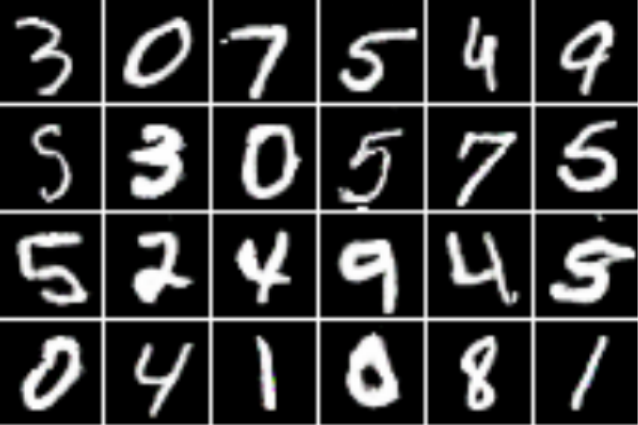}
    \caption{\acro{gan} samples.}
    \label{fig:mnist:igan:samp}
  \end{subfigure}
  \\
  \begin{subfigure}[c]{\textwidth}
    \includegraphics[width=\textwidth]{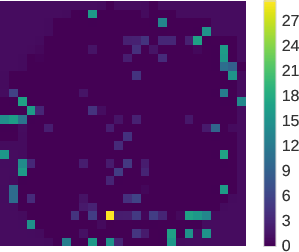}
    \caption{\acro{ard} weights.}
    \label{fig:mnist:igan:weights}
  \end{subfigure}
  \end{subfigure}
  ~
  \begin{subfigure}[c]{.7\textwidth}
    \includegraphics[width=\textwidth]{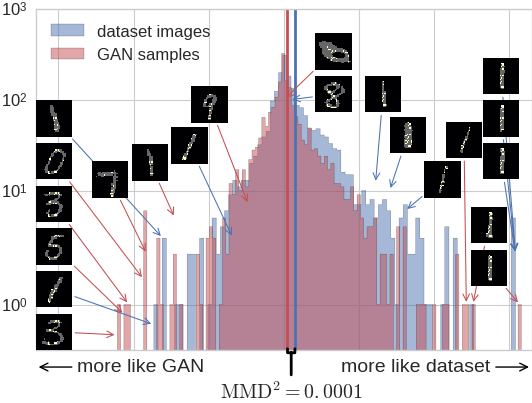}
    \caption{The witness function
    $\E_{x \sim P} k(x, \cdot) - \E_{y \sim Q} k(y, \cdot)$
    evaluated on a test set.
    Images are shown with \acro{ard} weights highlighted.
    Vertical lines show distribution means; the distance between them is small but highly consistent.}
    \label{fig:mnist:igan:witness}
  \end{subfigure}
  \caption{Model criticism of \citet{SalGooZarCheetal16}'s semi-supervised \acro{gan} on \acro{mnist}.}
\end{figure}

\paragraph{Model criticism}
As an example of a real-world two-sample testing problem,
we will consider distinguishing the output of a generative model from the reference distribution it attempts to reproduce.
We will use the semi-supervised \acro{gan} model of \citet{SalGooZarCheetal16},
trained on the \acro{mnist} dataset of handwritten images.\footnote{We used their code for a minibatch discrimination \acro{gan}, with 1000 labels, and chose the best of several runs.}
True samples from the dataset are shown in \cref{fig:mnist:true:samp};
samples from the learned model are in \cref{fig:mnist:igan:samp}.
\citet{SalGooZarCheetal16} called their results ``completely indistinguishable from dataset images,''
and reported that annotators on Mechanical Turk were able to distinguish samples only in 52.4\% of cases.
Comparing the results, however, there are several pixel-level artifacts that make distinguishing the datasets trivial;
our methods can pick up on these quickly.

To make the problem more interesting, we discretized the sampled pixels into black or white (which barely changes the images visually).
The samples are then in $\{0, 1\}^{28 \times 28}$.
We trained an automatic relevance determination (\acro{ard})-type kernel:
in the notation of \cref{sec:MMDandTestPower},
$z$ scales each pixel by some learned value,
and $k$ is a Gaussian \acro{rbf} kernel with a learned global bandwidth.
We optimized $\hat t$ on $2\,000$ samples in batches of size $500$
using the Adam optimizer \citep{adam},
where the learned weights are visualized in \cref{fig:mnist:igan:weights}.
This network has essentially perfect discriminative power:
testing it on 100 different samples with 1000 permutations for each test,
in 98 cases we obtained $p$-values of 0.000 and twice got $0.001$.
By contrast, using an \acro{rbf} kernel with a bandwidth optimized by maximizing the $t$ statistic gave a less powerful test:
the worst $p$-value in 100 repetitions was $0.135$, with power $57\%$ at the $\alpha = 0.01$ threshold.
An \acro{rbf} kernel based on the median heuristic, which here found a bandwidth five times the size of the $t$-statistic-optimized bandwidth, performed worse still: three out of 100 repetitions found a $p$-value of exactly $1.000$, and power at the $.01$ threshold was 42\%.
The learned weights show that the model differs from the true dataset along the outsides of images, as well as along a vertical line in the center.

We can investigate these results in further detail using the approach of \citet{LloGha15}, considering the witness function associated with the \acro{mmd},
which has largest amplitude where the probability mass of the two samples is most different.
Thus, samples falling at maxima and minima of the witness function best represent the difference in the distributions.
The value of the witness function on each sample is plotted in \cref{fig:mnist:igan:witness}, along with some images with different values of the witness function.
Apparently, the \acro{gan} is slightly overproducing images resembling the \texttt{/}-like digits on the left,
while underproducing vertical \texttt{1}s.
It is not the case that the \acro{gan} is simply underproducing \texttt{1}s in general:
the $p$-values of a $\chi^2$ contingency test between the outputs of digit classifiers on the two distributions are uniform.
This subtle difference in proportions among types of digits
would be quite difficult for human observers to detect.
Our testing framework allows the model developer to find such differences and decide whether to act on them.
One could use a more complex representation function $z$ to detect even more subtle differences between distributions.

\begin{figure}[]
  \centering
  \begin{subfigure}[c]{.23\textwidth}
    \includegraphics[width=\textwidth]{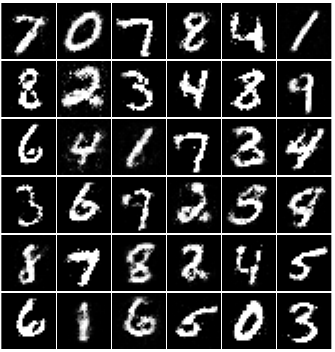}
    \caption{\acro{gmmn}}
    \label{fig:gmmn}
  \end{subfigure}
  ~
  \begin{subfigure}[c]{.23\textwidth}
    \includegraphics[width=\textwidth]{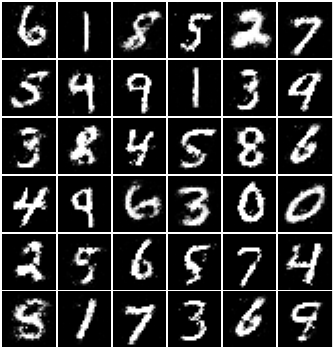}
    \caption{$t$-\acro{gmmn}}
    \label{fig:t-gmmn}
  \end{subfigure}
  ~
  \begin{subfigure}[c]{.23\textwidth}
    \includegraphics[width=\textwidth]{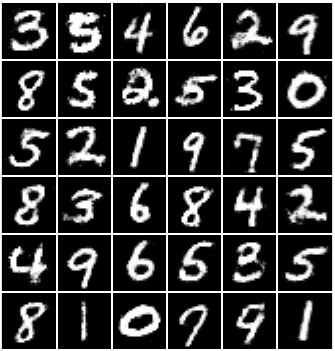}
    \caption{Feature matching}
    \label{fig:fm-mmd}
  \end{subfigure}
  ~
  \begin{subfigure}[c]{.23\textwidth}
    \includegraphics[width=\textwidth]{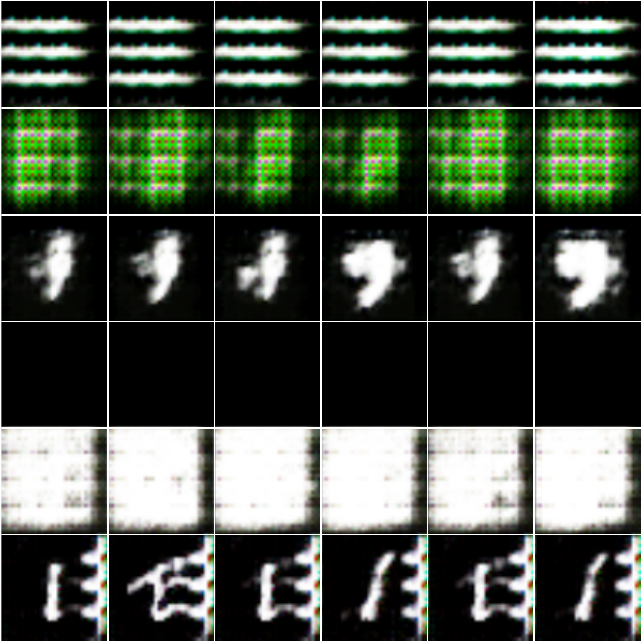}
    \caption{Improved \acro{gan}}
    \label{fig:igan-failure}
  \end{subfigure}
  \caption{\acro{MNIST} digits from various models. Part \subref{fig:igan-failure} shows six runs of the minibatch discrimination model of \citet{SalGooZarCheetal16}, trained without labels~---~the same model that, with labels, generated \cref{fig:mnist:igan:samp}. (The third row is the closest we got the model to generating digits without any labels.)}
    \label{fig:gan-samps}
\end{figure}

\paragraph{\acro{gan} criterion}
We now demonstrate the use of \acro{mmd} as a training criterion in \acro{gan}s.
We consider two basic approaches, and train on \acro{mnist}.\footnote{%
{\bf Implementation details:}
We used the architecture of \citet{li:gmmn}:
the generator consists of fully connected layers with sizes $10, 64, 256, 256, 1024, 784,$
each with ReLU activations except the last, which uses sigmoids.
The kernel function for \acro{gmmn}s is a sum of Gaussian \acro{rbf} kernels
with fixed bandwidths $2, 5, 10, 20, 40, 80$.
For the feature matching \acro{gan},
we use a discriminator with fully connected layers of size $512, 256, 256, 128, 64$, each with sigmoid activation.
We then concatenate the raw image and each layer of features as input to the same mixture of \acro{rbf} kernels as for \acro{gmmn}s.
We optimize with \acro{sgd}.
Initialization for all parameters are Gaussian with standard deviation $0.1$ for the
\acro{gmmn}s and $0.2$ for feature matching. Learning rates are $2, 0.02, 0.5$, respectively.
Learning rate for the feature matching discriminator is set to $0.01.$ All experiments are run for $50\,000$
iterations and use a momentum optimizer with with momentum $0.9$.}
First, the generative moment matching network (\acro{gmmn}; \cref{fig:gmmn}) approach \citep{li:gmmn,dziugaite:mmd-gan}
uses an  \acro{mmd} statistic computed with an \acro{rbf} kernel directly on the images
as the discriminator of a \acro{gan} model.
The $t$-\acro{gmmn} (\cref{fig:t-gmmn}) has the generator minimize the $\hat t_k$ statistic for a fixed kernel.\footnote{One could additionally update the kernel adversarially, by maximizing the $\hat t_k$ statistic based on generator samples, but we had difficulty in optimizing this model: the interplay between generator and discriminator adds some difficulty to this task.}
Compared to standard \acro{gmmn}s,
the $t$-\acro{gmmn} more directly attempts to make the distributions \emph{indistinguishable} under the kernel function;
it avoids a situation like that of \cref{fig:mnist:igan:witness},
where although the \acro{mmd} value is quite small, the two distributions are perfectly distinguishable due to the small variance.

Next, feature matching \acro{gan}s (\cref{fig:fm-mmd}) train the discriminator as a classifier like a normal \acro{gan},
but train the generator to minimize the \acro{mmd} between generator samples and reference samples with a kernel computed on intermediate features of the discriminator.
\citet{SalGooZarCheetal16} proposed feature matching using the mean  features at the top of the discriminator (effectively using an \acro{mmd} with a linear kernel);
we instead use \acro{mmd} with a mixture of \acro{rbf} kernels,
ensuring that the full feature distributions match, rather than just their means.
This helps avoid the common failure mode of \acro{gan}s where the generator collapses to outputting a small number of samples considered highly realistic by the discriminator.
Using the \acro{mmd}-based approach, however, no single point can approximate the feature {\em distribution}.
The minibatch discrimination approach of \citet{SalGooZarCheetal16} attempts to solve the same problem,
by introducing features measuring the similarity of each sample to a selection of other samples,
but we were unable to get it to work without labels to force the discriminator in a reasonable direction;
\cref{fig:igan-failure} demonstrates some of those failures,
with each row showing six samples from each of six representative runs of the model.

\subsubsection*{Acknowledgements}
We would like to thank Tim Salimans, Ian Goodfellow, and Wojciech Zaremba for providing their code and for gracious assistance in using it, as well as Jeff Schneider for helpful discussions.

\bibliography{iclr2017_conference}
\bibliographystyle{iclr2017_conference}

\appendix
\section{Variance of the Pairwise MMD Estimator} \label{sec:variance}

The publication version of this appendix contained some small mistakes.
Please refer instead to \citet{unbiased-var-ests};
in particular, the estimator \eqref{eq:vhat-m} is equivalent to (4) of that document.

\end{document}